\pdfoutput=1

\documentclass[11pt]{article}

\usepackage[preprint]{acl}

\usepackage{times}
\usepackage{latexsym}
\usepackage{booktabs}
\usepackage[T1]{fontenc}

\usepackage[utf8]{inputenc}

\usepackage{microtype}
\usepackage{tcolorbox}
\usepackage{inconsolata}

\usepackage{graphicx}

\def\retrievaltextcircled#1{\large{\textcircled{\small{\color{blue}#1}}}\normalsize}
\def\reranktextcircled#1{\large{\textcircled{\small{\textcolor{red}{#1}}}}\normalsize}
\def\datatextcircled#1{\large{\textcircled{\small{#1}}}\normalsize}
\def\fusiontextcircled#1{\large{\textcircled{\small{\color{green}#1}}}\normalsize}

\def\doctextcircled#1{\large{\textcircled{\small{\color{purple}#1}}}\normalsize}

\title{EasyRAG: Efficient Retrieval-Augmented Generation Framework for Automated Network Operations}

\author{
    Zhangchi Feng\textsuperscript{\rm 1}, Dongdong Kuang\textsuperscript{\rm 1}, Zhongyuan Wang\textsuperscript{\rm 1},\\\bf {Zhijie Nie\textsuperscript{\rm 1}, Yaowei Zheng\textsuperscript{\rm 1}}, Richong Zhang\textsuperscript{\rm 1}\thanks{\ \ Corresponding author}\\
    \textsuperscript{\rm 1}CCSE, School of Computer Science and Engineering, Beihang University, Beijing, China\\
    \texttt{\{zcmuller,kuangdd,wangzy23,hiyouga\}@buaa.edu.cn}, \texttt{\{niezj,zhangrc\}@act.buaa.edu.cn}
}

\begin{document}
\maketitle
\begin{abstract}
This paper presents EasyRAG, a simple, lightweight, and efficient retrieval-augmented generation framework for automated network operations\footnote{This work is a technical report of our solution at the 2024 (7th) CCF International AIOps Challenge. The official website of the Challenge is \url{https://competition.aiops-challenge.com}}. Our framework has three advantages. The first is \textbf{accurate question answering}. We designed a straightforward RAG scheme based on (1) a specific data processing workflow (2) dual-route sparse retrieval for coarse ranking (3) LLM Reranker for reranking (4) LLM answer generation and optimization. This approach achieved first place in the GLM4 track in the preliminary round and second place in the GLM4 track in the semifinals. The second is \textbf{simple deployment}. Our method primarily consists of BM25 retrieval and BGE-reranker reranking, requiring no fine-tuning of any models, occupying minimal VRAM, easy to deploy, and highly scalable; we provide a flexible code library with various search and generation strategies, facilitating custom process implementation. The last one is \textbf{efficient inference}. We designed an efficient inference acceleration scheme for the entire coarse ranking, reranking, and generation process that significantly reduces the inference latency of RAG while maintaining a good level of accuracy; each acceleration scheme can be plug-and-play into any component of the RAG process, consistently enhancing the efficiency of the RAG system. Our code and data are released at \url{https://github.com/BUAADreamer/EasyRAG}.

\end{abstract}

\section{Introduction}
\label{sec:method}
\begin{figure*}[htp!]
    \centering
    \includegraphics[width=\linewidth]{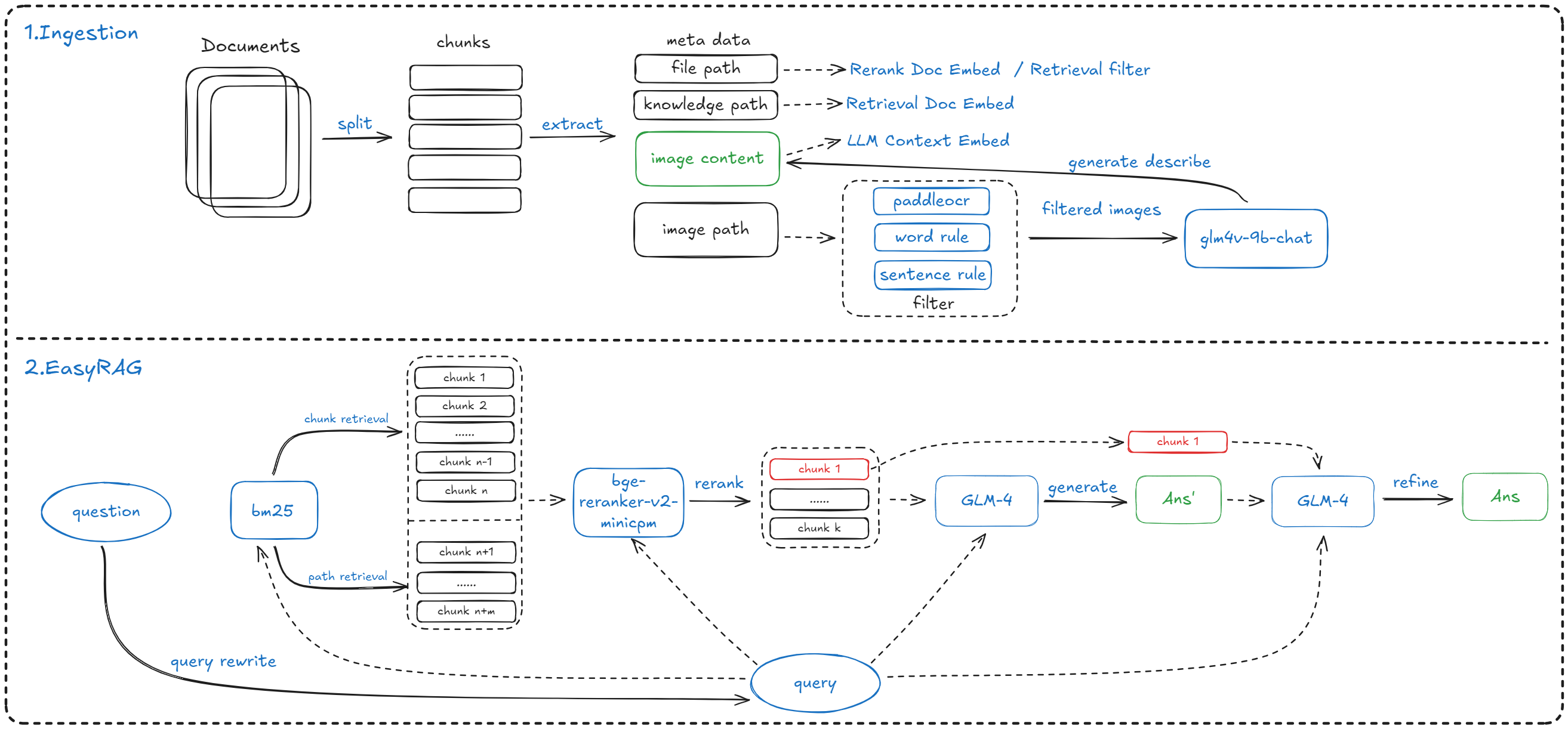}
    \caption{EasyRAG Framework}
    \label{fig:overview}
\end{figure*}

Our solution can be summarized by Fig.~\ref{fig:overview}, which includes a data processing workflow (Section~\ref{sec:ingestion}) and the RAG process (Section~\ref{sec:rag-pipeline}).

\subsection{Ingestion}\label{sec:ingestion}
\subsubsection{zedx file processing}
Due to the discovery that the original processing script missed some files, we have reprocessed the zedx files using the following steps:

\begin{enumerate}
    \item \textbf{zedx Decompression}: Decompress the official source data from four .zedx files, obtaining four packages of HTML documents.
    \item \textbf{Path Parsing}: Read the \textbf{knowledge path} and the actual \textbf{file path} from the nodetree.xml in each document package.
    \item \textbf{Document Extraction}: Extract the text, image titles, and image paths from each HTML document using BeautifulSoup.
    \item \textbf{Saving}: Save the document text in txt format, maintaining the relative location consistent with the HTML document. Also, save the knowledge path, file path, and image path information.
\end{enumerate}

\subsubsection{Text Segmentation}
\paragraph{Segmentation Settings} We used SentenceSplitter for document segmentation, initially splitting into sentences using Chinese punctuation, then merging according to the set text block size. The used block size (chunk-size) is 1024, and the block overlap size (chunk-overlap) is 200.

\paragraph{Eliminating Path Influence in Segmentation} In practice, we found that the original implementation of llama-index used a simple but unstable method of handling path information, subtracting the file path length from the text length to determine the actual text length used. This approach could cause different segmentation results with the same chunk-size and chunk-overlap, depending on the data path. During the preliminary competition, we observed that changing paths could lead to a fluctuation of up to 3 percentage points in the final evaluation results, which is obviously unacceptable in practice. To address this issue, we implemented a custom segmentation class that eliminates the use of path length, thereby ensuring stable reproducibility.

\subsubsection{Image Information Extraction}
\paragraph{Image Content Extraction Using a Multimodal Large Model} First, we extracted information from all images using GLM-4V-9B~\cite{glm2024chatglm}. We found that the following simple prompt achieves good results:
\begin{tcolorbox}\small
Briefly describe the image
\end{tcolorbox}

\paragraph{Image Filtering Based on Various Rules} We found that a small number of images are beneficial for the final question answering, but not all images are useful. Therefore, we designed a flexible strategy to filter out useless images using the following steps:
\begin{enumerate}
    \item Use the PP-OCRv4 model\footnote{\href{https://github.com/PaddlePaddle/PaddleOCR}{https://github.com/PaddlePaddle/PaddleOCR}} to extract text content from images and filter out images that do not contain Chinese.
    \item Filter images whose titles contain specific keywords (e.g., network diagrams, architecture).
    \item Filter images that are referenced in the text in a specific way (e.g., configuration as shown in Figure x, file as shown in Figure x).
\end{enumerate}
With these filtering steps, we reduced the number of images from an original 6000 to fewer than 200. Notably, the filtering process is easily configurable, allowing for tuning to suit real-world scenarios.

\subsection{RAG Pipeline}\label{sec:rag-pipeline}

\subsubsection{Query Rewriting}
\label{subsec:query_edit}
During the competition, given that the queries were very brief and we identified issues with some queries being semantically awkward or having unclear keywords. For instance, "What types of alarms are there in EMSPLuS?" and "What are the sources of faults?". Before inputting these queries into the RAG Pipeline, we used a Large Language Model (LLM, GLM4) for query rewriting, which involved two methods: query expansion and Hypothetical Document Embedding (HyDE) \citep{gao2022precise}.

\paragraph{Query Expansion}
During the preliminary round, we summarized the characteristics of queries in the current operational maintenance scenario:
\begin{itemize}
    \item Technical keywords in queries are crucial.
    \item Queries are short and vary greatly in the amount of information provided.
\end{itemize}
In this context, we attempted to summarize the key terms in the queries or other potentially relevant keywords using the LLM, i.e., using the LLM's knowledge for keyword association and summary in the fields of operation and communication. This is referred to as \textbf{keyword expansion}.

After manually annotating several data points with keywords and potential associations, we utilized the LLM (GLM4) for few-shot keyword summarization and expansion. Following \citep{wang2023query2doc}, we generated new queries by directly concatenating the expanded keywords with the original query and then re-summarizing them using a large language model.

Let $\mathcal{L}$ represent the Large Language Model LLM, with $q$ and $p$ denoting the initial query and the prompt, respectively. $p_{exp}$ represents the expanded query prompt, including manually annotated data points, and $p_{sum}$ represents the prompt for summarizing and concatenating the sentence and expanded keywords using the large model.

\paragraph{HyDE} In situations where queries lack specificity or identifiable elements, making it difficult for both dense and sparse retrieval methods to locate the target document, we designed a set of hypothetical document embedding methods, inspired by \cite{gao2022precise}.

\begin{figure}
    \centering
    \includegraphics[width=0.9\linewidth]{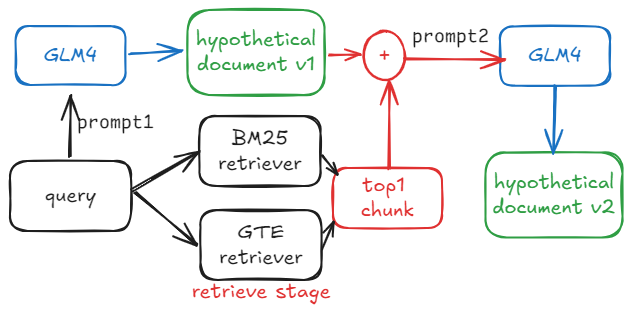}
    \caption{Process of generating hypothetical documents}
    \label{fig:hyde}
\end{figure}

For the generation of fictional documents, we devised two approaches, as shown in Figure~\ref{fig:hyde}. Initially, following the paper's methodology, we input the prompt $p_{hy}$ and the original question $q$ into the large language model $\mathcal{L}$ to produce the fictional document $q'_0$. However, during the semifinals, we discovered that such fictional documents contained a significant amount of irrelevant keywords and redundant information due to the large model's hallucinations, greatly affecting the effectiveness of the retrieval process. Therefore, we attempted to minimize the hallucinations and redundant information in the initial fictional document $q'_0$ by using the BM25 algorithm and dense retrieval (using GTE-QWEN encoding) to identify the most relevant top1 document and use it for context prompting.

For the generated fictional documents, we also adopted two application methods: 1. Using the fictional document $q'$ combined with the original document $q$ for \textbf{coarse ranking} retrieval. 2. Using only the fictional document $q'$ combined with the original document $q$ for \textbf{re-ranking} of retrieval results.

\subsubsection{Dual-route Sparse Retrieval for Coarse Ranking}
In the sparse retrieval section, we utilized the BM25 algorithm to construct the retriever. The core idea of BM25 is based on term frequency (TF) and inverse document frequency (IDF), and it also incorporates document length information to calculate the relevance between the document and query $q$. Specifically, the BM25 retriever primarily consists of a Chinese tokenizer and a stopword list. We will introduce each component in detail.

\paragraph{Chinese Tokenizer}
For the Chinese tokenizer, we used the widely known jieba Chinese tokenizer\footnote{\href{https://github.com/fxsjy/jieba}{https://github.com/fxsjy/jieba}}, which is lightweight and supports multi-threaded mode to accelerate tokenization and part-of-speech analysis. It also allows for customization of word frequency or dictionaries to adjust tokenization preferences. For the tokenizer, we also attempted to customize the vocabulary; in the current 5G communication maintenance scenario, we chose a related IT field lexicon collected by Tsinghua University\footnote{\href{http://thuocl.thunlp.org}{http://thuocl.thunlp.org}} loaded into the tokenizer. However, the results in practice were mediocre, so we ultimately continued using the original jieba lexicon.

\paragraph{Stopword List}
For the Chinese stopword list, we adopted the common Chinese stopword list collected by Harbin Institute of Technology\footnote{\href{https://github.com/goto456/stopwords}{https://github.com/goto456/stopwords}} as a reference for filtering out meaningless words during Chinese tokenization. By filtering out irrelevant words and special symbols, we improve the hit rate of valid keywords and increase the recall rate of correct documents.

\paragraph{Dual-route Retrieval}
The BM25 dual-route retrieval for coarse ranking consists of text block retrieval and path retrieval.
\begin{enumerate}
    \item Text block retrieval. Use BM25 to search the segmented text blocks, recalling the top 192 text blocks with a coarse ranking score greater than 0.
    \item Path retrieval. Considering that some questions are highly relevant to our extracted knowledge paths, such as the question "How many types of VNF elasticity are there?", where both VNF and elasticity can be directly found in related knowledge paths. Hence, we designed a path search using BM25 to search the knowledge paths, recalling the top 6 text blocks with a coarse ranking score greater than 0.
\end{enumerate}

\paragraph{Retrieval Process}
The BM25 retriever follows the document retrieval process below for a given query $q$:
\begin{enumerate}
    \item Document Expansion. For text block retrieval, we concatenate the file path and each text block together to serve as expanded documents for retrieval.
    \item Document Preprocessing. First, filter all documents (text blocks or paths) with stopwords, then use the Chinese tokenizer for tokenization, and pre-compute the IDF scores of the documents.
    \item Query Processing. Filter the query $q$ with stopwords and perform Chinese tokenization.
    \item Similarity Recall. Count the keywords of query $q$ and calculate the TF values of each document, compute the relevance scores based on TF and IDF values, and recall relevant documents based on scores.
    \item File Path Filtering. For text block retrieval, we use the file paths provided in the competition to compare metadata, filtering out text blocks from other sources.
\end{enumerate}

\subsubsection{Dense Retrieval for Coarse Ranking}

In the dense retrieval section, we employed the gte-Qwen2-7B-instruct model developed by Alibaba\footnote{\href{https://huggingface.co/Alibaba-NLP/gte-Qwen2-7B-instruct}{https://huggingface.co/Alibaba-NLP/gte-Qwen2-7B-instruct}}~\cite{li2023towards}, which has achieved advanced results on the MTEB benchmark.

\paragraph{Retrieval Process} The dense retriever for a given query $q$ follows the specific document retrieval process as outlined below:
\begin{enumerate}
    \item Document Expansion. We concatenate the file path with each text block to serve as expanded documents for retrieval.
    \item Document Encoding. All text blocks are input into the model to be encoded and the representations are stored in a Qdrant\footnote{\href{https://qdrant.tech/}{https://qdrant.tech/}} vector database.
    \item Query Encoding. Using a query prompt template, we transform $q$ into an input suitable for the GTE model and encode it using the model.
    \item Similarity Recall. During retrieval, cosine similarity is used for matching, recalling the top 288 text blocks.
    \item File Path Filtering. Using the file paths provided in the competition, we employ a Qdrant filter to eliminate text blocks from other sources.
\end{enumerate}

\subsubsection{LLM Reranker Re-ranking}

We utilized the bge-reranker-v2-minicpm-layerwise model~\cite{chen2024bge}, a LLM Reranker trained on a hybrid of multiple multilingual ranking datasets using MiniCPM-2B-dpo-bf16. This model exhibits advanced ranking performance in both Chinese and English and includes accompanying tool code, which can be conveniently fine-tuned for specific scenarios.

\paragraph{Re-ranking Process} The LLM-Reranker for a given query $q$ and $k'$ coarsely ranked text blocks follows the specific document ranking process as outlined below:
\begin{enumerate}
    \item Document Expansion. We concatenate the knowledge paths with each text block to serve as expanded documents for retrieval.
    \item Text Processing. Combine $q$ with the $k'$ text blocks to form $k'$ query-document pairs, which are then input into the tokenizer to generate input data for the LLM.
    \item Similarity Ranking. The input data is fed into the LLM to obtain re-ranking scores for the query and each text block, and the blocks are sorted according to these scores. The highest ranked $k$ (typically 6) text blocks are returned.
\end{enumerate}

\subsubsection{Multi-route Ranking Fusion}
\paragraph{Fusion Algorithm} Since we designed multiple routes for coarse retrieval, it is also necessary to design corresponding ranking fusion strategies. We primarily used two strategies: simple merging and Reciprocal Rank Fusion (RRF). The simple merging strategy directly de-duplicates and merges text blocks obtained from multiple routes. Reciprocal Rank Fusion sums the reciprocals of the ranks of the same document across multiple retrieval paths to compute the fusion score for re-ranking.

\paragraph{Coarse Ranking Fusion} The most straightforward use of ranking fusion is to merge the text blocks obtained from multi-route coarse retrieval into a single set of text blocks, which are then passed to the Reranker for re-ranking. In the semifinals, we used simple merging to combine results from two sparse retrieval routes.

\paragraph{Re-ranking Fusion} We can also perform fusion after coarse ranking and re-ranking for each route. In the preliminary rounds, we fused text blocks from sparse and dense retrieval routes. For these two routes, we designed three re-ranking fusion methods. (1) Use RRF to merge the results after coarse and fine ranking.\label{item:fusion0} (2) Input the text blocks from each route into the LLM to obtain respective answers, selecting the longer answer as the final one.\label{item:fusion1} (3) Input the text blocks from each route into the LLM to obtain respective answers and directly concatenate the answers from all routes.\label{item:fusion2}

\subsubsection{LLM Answer Generation}\label{sec:generate}
In this section, we first concatenate the contents of the top 6 text blocks obtained from re-ranking using the following template to create a context string:
\begin{tcolorbox}
    \#\#\# Document 0: \{{\color{blue}chunk\_i}\}\\
    ...\\
    \#\#\# Document 5: \{{\color{blue}chunk\_i}\}\\
\end{tcolorbox}
Note that the text blocks input into GLM4 here include concatenated image content, whereas the text blocks in the previous coarse and re-ranking processes did not include image content.

We then combine the context string and the question using the following question-and-answer template, and input it into GLM4 to obtain an answer:
\begin{tcolorbox}
The context information is as follows:

-\--\--\--\--\--\--\--\--\--\\

\{{\color{blue}context\_str}\}

-\--\--\--\--\--\--\--\--\--\\

Please answer the following question based on the context information and not your own knowledge. Answers can be itemized. If the context does not contain relevant information, you may respond with "uncertain" and should not restate the context information:\\

\{{\color{blue}query\_str}\}\\

Answer:
\end{tcolorbox}

Additionally, we have designed other formats of question-and-answer templates. Inspired by Chain-of-Thought~\cite{wei2022chain}, we designed a Chain-of-Thought question-and-answer template (see Appendix \ref{sec:cot_temp}). Drawing from COSTAR~\cite{costar}, we designed a markdown format question-and-answer template (see Appendix \ref{sec:md_temp}). To emphasize the importance of the top1 document, we designed a focused question-and-answer template (see Appendix \ref{sec:lean_temp}). Related experimental results are discussed therein.

\subsubsection{LLM Answer Optimization}
Due to our observation that the LLM gives attention to each text block, which may result in the effective information from the top1 text block not being fully utilized, we designed an answer integration prompt (see Appendix \ref{sec:merge_temp}). This prompt allows us to integrate and supplement the answers derived from the 6 text blocks using the top1 text block, leading to the final answer.

\begin{table*}[ht!]
\setlength{\tabcolsep}{13pt}
\centering
\begin{tabular}{ccccccc}
\toprule
\textbf{id} & \textbf{data} & \textbf{chunk} & \textbf{coarse ranking} & \textbf{re-ranking} & \textbf{fusion} & \textbf{accuracy} \\
\midrule
0 & \datatextcircled{0} & 1024,50  & \retrievaltextcircled{0},3 & - & - & 57.86 \\
1 & \datatextcircled{0} & 1024,50 & \retrievaltextcircled{1},8 & - & - & \textbf{68.59} \\
2 & \datatextcircled{0} & 1024,50 & \retrievaltextcircled{3},8 & - & - & \textbf{69.55} \\
\midrule
3 & \datatextcircled{0} & 1024,50 & \retrievaltextcircled{1},192 & \reranktextcircled{0},8 & - & \textbf{73.73} \\
4 & \datatextcircled{0} & 1024,50 & \retrievaltextcircled{1},256 & \reranktextcircled{0},8 & - & 70.68 \\
5 & \datatextcircled{0} & 1024,50 & \retrievaltextcircled{2},192 & \reranktextcircled{1},8 & - & 69.25 \\
\midrule
6 & \datatextcircled{0} & 1024,50 & \retrievaltextcircled{1},288 & \reranktextcircled{2},8 & - & 77.07 \\
7 & \datatextcircled{1} & 1024,50 & \retrievaltextcircled{1},288 &\reranktextcircled{2},8 & - & 77.51 \\
8 & \datatextcircled{2} & 1024,50 & \retrievaltextcircled{1},288 &\reranktextcircled{2},8 & - & 77.92 \\
9 & \datatextcircled{2} & 1024,50 & \retrievaltextcircled{1},256 &\reranktextcircled{2},8 & - & 78.49 \\
10 & \datatextcircled{1} & 1024,50 & \retrievaltextcircled{3},192 &\reranktextcircled{2},6 & - & 80.90 \\
11 & \datatextcircled{2} & 1024,50 & \retrievaltextcircled{3},192 &\reranktextcircled{2},6 & - & 81.38 \\
12 & \datatextcircled{2} & 1024,100 & \retrievaltextcircled{3},192 &\reranktextcircled{2},6 & - & 81.77 \\
13 & \datatextcircled{2} & 1024,100 & \retrievaltextcircled{3},192 &\reranktextcircled{3},6 & - & 81.88 \\
14 & \datatextcircled{2} & 1024,200 & \retrievaltextcircled{3},192 &\reranktextcircled{2},6 & - & 82.87 \\
15 & \datatextcircled{2} & 1024,200 & \retrievaltextcircled{3},192 &\reranktextcircled{3},6 & - & \textbf{82.97} \\
16 & \datatextcircled{2} & 1024,200 & \retrievaltextcircled{4},288 &\reranktextcircled{3},6 & - & \textbf{83.02} \\
\midrule
17 & \datatextcircled{2} & 1024,200 & \retrievaltextcircled{4},288 \retrievaltextcircled{3},192 &\reranktextcircled{3},6 & \fusiontextcircled{0} & 81.80 \\
18 & \datatextcircled{2} & 1024,200 & \retrievaltextcircled{4},288 \retrievaltextcircled{3},192 &\reranktextcircled{3},6 & \fusiontextcircled{1} & 82.50 \\
19 & \datatextcircled{2} & 1024,200 & \retrievaltextcircled{4},288 \retrievaltextcircled{3},192 &\reranktextcircled{3},6 & \fusiontextcircled{2} & \textbf{83.45} \\
20 & \datatextcircled{2} & 1024,200 & \retrievaltextcircled{4},288 \retrievaltextcircled{3},192 &\reranktextcircled{3},6 & \fusiontextcircled{3} & \textbf{83.70} \\
21 & \datatextcircled{2} & 1024,200 & \retrievaltextcircled{4},288 \retrievaltextcircled{3},192 &\reranktextcircled{3},6 & \fusiontextcircled{4} & \textbf{84.38} \\
\bottomrule
\end{tabular}
\caption{Preliminary round experimental results. In the 'Chunk' column, the two numbers represent chunk\_size and chunk\_overlap, respectively. In the 'Coarse Ranking' and 'Re-ranking' columns, multiple search paths are separated by spaces, and within each search path, the components separated by commas represent the retrieval/sorting method and top-k, respectively.}
\label{tab:main_res1}
\end{table*}

\begin{table*}[ht]
\centering
\begin{tabular}{ccccccccc}
\toprule
\textbf{id} & \textbf{data} & \textbf{chunk} & \textbf{coarse ranking} & \textbf{re-ranking} & \textbf{fusion} & \textbf{image} & \textbf{answer merge} & \textbf{accuracy} \\
\midrule
0 & \datatextcircled{2} & 960,200 & \retrievaltextcircled{3},192 &\reranktextcircled{3},6 & -& - & - & \textbf{91.53} \\
1 & \datatextcircled{2} & 960,200 & \retrievaltextcircled{4},288 &\reranktextcircled{3},6 & -& - & - & 88.40 \\
2 & \datatextcircled{2} & 960,200 & \retrievaltextcircled{4},288 \retrievaltextcircled{3},192 &\reranktextcircled{3},6 & \fusiontextcircled{2} & -& -  & 90.00 \\
3 & \datatextcircled{3} & 1024,200 & \retrievaltextcircled{3},192 &\reranktextcircled{3},6 & - & -& -  & 90.26 \\
4 & \datatextcircled{4} & 1024,200 & \retrievaltextcircled{3},192 &\reranktextcircled{3},6 & - & -& - & \textbf{91.38} \\
\midrule
5 & \datatextcircled{3} & 1024,200 & \retrievaltextcircled{3},192,\doctextcircled{1} &\reranktextcircled{3},6 & - & -& -  &\textbf{92.70} \\
6 & \datatextcircled{3} & 1024,200 & \retrievaltextcircled{3},192,\doctextcircled{1} &\reranktextcircled{3},6,\doctextcircled{1} & -& -& -  & 89.30 \\
7 & \datatextcircled{3} & 1024,200 & \retrievaltextcircled{3},192,\doctextcircled{1} &\reranktextcircled{3},6,\doctextcircled{2} & -& -& -  & 87.12 \\
8 & \datatextcircled{3} & 1024,200 & \retrievaltextcircled{3},192,\doctextcircled{2} &\reranktextcircled{3},6 & -& -& -  & 92.43 \\
9 & \datatextcircled{3} & 1024,200 & \retrievaltextcircled{3},192,\doctextcircled{2} &\reranktextcircled{3},6,\doctextcircled{1} & -& -& -  & \textbf{93.11} \\
10 & \datatextcircled{3} & 1024,200 & \retrievaltextcircled{3},192,\doctextcircled{2} &\reranktextcircled{3},6,\doctextcircled{2} & -& -& -  & 90.17 \\
\midrule
11 & \datatextcircled{3} & 1024,200 & \retrievaltextcircled{3},192,\doctextcircled{2} &\reranktextcircled{3},6,\doctextcircled{1} & -& OCR Filter & -  & 92.5 \\
12 & \datatextcircled{3} & 1024,200 & \retrievaltextcircled{3},192,\doctextcircled{2} &\reranktextcircled{3},6,\doctextcircled{1} & -& Rule Filter & -  & \textbf{94.24} \\
13 & \datatextcircled{3} & 1024,200 & \retrievaltextcircled{3},192,\doctextcircled{2} \retrievaltextcircled{5},6&\reranktextcircled{3},6,\doctextcircled{1} & \fusiontextcircled{0} & Rule Filter & -  & \textbf{94.49} \\
\midrule
14 & \datatextcircled{3} & 1024,200 & \retrievaltextcircled{3},192,\doctextcircled{2} \retrievaltextcircled{5},6&\reranktextcircled{3},6,\doctextcircled{1} & \fusiontextcircled{0}& Rule Filter & document concat  & \textbf{96.65} \\
15 & \datatextcircled{3} & 1024,200 & \retrievaltextcircled{3},192,\doctextcircled{2} \retrievaltextcircled{5},6&\reranktextcircled{3},6,\doctextcircled{1} & \fusiontextcircled{0}& Rule Filter & prompt merge  & \textbf{95.72} \\
\bottomrule
\end{tabular}
\caption{Semi-final experimental results. In the 'Chunk' column, the two numbers represent chunk\_size and chunk\_overlap, respectively. In the 'Coarse Ranking' and 'Re-ranking' columns, multiple search paths are separated by spaces, and within each search path, the components separated by commas represent the retrieval/sorting method, top-k, and the type of document expansion (if no expansion is applied, it is not listed).}
 \label{tab:main_res2}

\end{table*}

\section{Accuracy}

\subsection{Abbreviations Introduction}
For ease of writing, we first introduce some important component identifiers.

\paragraph{Data} \datatextcircled{0} represents the official processed txt data. \datatextcircled{1} represents our own processed version 0 txt data, which supplements some missing data compared to the official data. \datatextcircled{2} is similar to \datatextcircled{1}, but each txt begins with a concatenated knowledge path. \datatextcircled{3} represents our own processed version 1 txt data, which retains more markdown-structured data consistent with the official data compared to version 0. \datatextcircled{4} is similar to \datatextcircled{3} but begins each txt with a concatenated knowledge path.

\paragraph{Coarse Ranking} \retrievaltextcircled{0} represents bge-small-zh-v1.5, \retrievaltextcircled{1} represents bge-base-zh-v1.5, \retrievaltextcircled{2} represents bce-embedding-base\_v1, \retrievaltextcircled{3} represents bm25 text block retrieval, \retrievaltextcircled{4} represents gte-Qwen2-7B-instruct, \retrievaltextcircled{5} represents bm25 knowledge path retrieval.

\paragraph{Re-ranking} \reranktextcircled{0} represents bge-reranker-v2-m3, \reranktextcircled{1} represents bce-reranker-base\_v1, \reranktextcircled{2} represents the 40-layer bge-reranker-v2-minicpm-layerwise, \reranktextcircled{3} represents the 28-layer bge-reranker-v2-minicpm-layerwise.

\paragraph{Fusion} \fusiontextcircled{0} represents simple merging for coarse ranking, \fusiontextcircled{1} represents RRF fusion for coarse ranking, \fusiontextcircled{2} represents re-ranking fusion using method~\ref{item:fusion0}, \fusiontextcircled{3} represents re-ranking fusion using method~\ref{item:fusion1}, \fusiontextcircled{4} represents re-ranking fusion using method~\ref{item:fusion2}.

\paragraph{Document Expansion} In the Coarse Rank and Re-rank columns, \doctextcircled{1} indicates that the document concatenates the file path, \doctextcircled{2} indicates that the document concatenates the knowledge path.

\subsection{Preliminary Experiments}

In the preliminary round, our main results are displayed in Table~\ref{tab:main_res1}, and improvements were made in the following four stages:
\begin{enumerate}
    \item Single-route coarse retrieval (0-2). We explored the retrieval effects of the bge-zh-v1.5 series (small, base, large) and bm25. We found that bge-base-zh-v1.5 and bm25 performed best when the top 8 results were taken; too many or too few results could lead to inaccuracies in LLM comprehension or a lack of necessary information.
    \item Re-ranking based on BERT-Reranker (3-5). We explored the effects of BERT-based bge series rerankers and bce-reranker-base\_v1, finding that bge-reranker-v2-m3 performed best. The best results for coarse ranking were between 192-288, and re-ranking performed best at top 8.
    \item Re-ranking based on LLM-Reranker (6-16). In this part, we made the following explorations: (1) We explored the effects of LLM-based bge series rerankers, finding that bge-reranker-v2-minicpm-layerwise significantly outperformed BERT-based rerankers, bringing an improvement of more than 3 percentage points. (2) We explored the dense retrieval effects of LLM-based embedding models, finding that gte-Qwen2-7B-Instruct, due to longer context lengths and a larger model size, performed better than the bge-zh-v1.5 series. (3) We perfected the data processing workflow, supplementing missing data from the official process, which brought a 1 percentage point improvement. (4) We optimized chunk parameters, finding that increasing chunk-overlap to preserve more complete semantic information brought a 2 percentage point improvement.
    \item Dual-route sorting fusion (17-21). In this part, we explored the impact of different sorting fusion strategies for sparse and dense routes on the results. Among them, coarse ranking fusion performed lower than the results of both routes. In re-ranking fusion, all three methods achieved higher results than the individual routes. The operations of answer concatenation and taking the longer answer, due to the need to generate an answer separately for each route before fusion, are less efficient and unstable. Therefore, in general practice, the first type of re-ranking fusion strategy is preferred, i.e., RRF fusion of the results reranked separately for each route, serving as the final context input into the LLM.
\end{enumerate}

\subsection{Semi-final Experiments}
In the preliminary round, we displayed our main results in Table~\ref{tab:main_res2}, and we made improvements through the following four stages:
\begin{enumerate}
    \item Exploration of Coarse Ranking Schemes (0-4). We optimized the data processing from the preliminary round, preserving more structured semantic information, and explored some of the better strategies from the preliminary round. We found that dense retrieval performed poorly, while BM25 alone could achieve good results.
    \item Document Extension for Sorting (5-10). We explored the impact of appending path strings to each text block during coarse and fine ranking. We found that adding paths during coarse ranking brought significant gains, and appending file paths during fine ranking provided certain benefits. We ultimately selected a document extension scheme where knowledge paths were inserted during coarse ranking and file paths during fine ranking. This part brought about a 2\% improvement.
    \item Utilization of Image Information (11-13). In this part, we explored the use of image information and found that the number of images after coarse screening with OCR in Chinese was large and varied. After fine screening with rules, we achieved a 1\% improvement over previous methods.
    \item Answer Optimization (14-15). We discovered that concatenating the top1 text block with the answer could lead to a 2\% improvement. Considering its practical effectiveness, we designed answer integration prompts that allow the LLM to supplement and optimize the answer in conjunction with the top1 text block, improving performance by about 2\%.
\end{enumerate}

\subsection{Exploratory Experiments}
\subsubsection{Query Rewriting}
For the query expansion and HyDE methods mentioned in Section~\ref{subsec:query_edit}, we tested them during both the preliminary and semi-final stages, with results displayed in Tables~\ref{tab:query_expansion} and \ref{tab:hyde}, respectively. Overall, since the query terms in the preliminary and semi-final competitions were already relatively specific, query rewriting did not bring any benefit. These rewriting methods might be more effective when user queries are incomplete.

\begin{table}[]
\small
    \centering
    \begin{tabular}{c|c}
    \toprule
        \textbf{Method} &\textbf{Preliminary Accuracy} \\
        \midrule
        Original & \textbf{82.0} \\
        Concat & 78.2\\
        Summary & 79.4 \\
        \bottomrule
    \end{tabular}
    \caption{Rewrite Performance}
    \label{tab:query_expansion}
\end{table}

\begin{table}[]
\small
    \centering
    \begin{tabular}{c|c}
    \toprule
        \textbf{Method} &\textbf{Semifinal Accuracy} \\
        \midrule
        Original& \textbf{92.7} \\
        Retrieval+HyDE& 89.2\\
        rerank+HyDE& 88.2 \\
        \bottomrule
    \end{tabular}
    \caption{HyDE Performance}
    \label{tab:hyde}
\end{table}

\subsubsection{Prompt Types}
We tested different prompt types mentioned in Section~\ref{sec:generate} during the semi-final stage, and the results are shown in Table~\ref{tab:prompt}. We found that the best results were still achieved with simple question-and-answer prompts. The Chain-of-Thought question-and-answer template led to too much explanatory output, the Markdown format question-and-answer template resulted in some extraneous characters, and the focused question-and-answer template did not significantly differ from the original template. Furthermore, through extensive experimentation with more prompts, we discovered that the GLM4 performs better with simpler prompts; more complex, structured prompts tend to have a negative impact.

\begin{table}[]
\small
    \centering
    \begin{tabular}{c|c}
    \toprule
        \textbf{Prompt Type} & \textbf{Semi-final Accuracy} \\
        \midrule
        Normal QA Template & \textbf{94.49} \\
        CoT QA Template & 89.75 \\
        Markdown Format QA Template & 92.27 \\
        Focused QA Template & 93.51 \\
        \bottomrule
    \end{tabular}
    \caption{Effects of Different Prompts}
    \label{tab:prompt}
\end{table}

\section{Resource Consumption}

In our RAG process, only the Reranker requires significant GPU memory consumption. Thanks to enabling bfloat16, model loading requires only 5GB of GPU memory\footnote{We also experimented with \textbf{8-bit quantization} and \textbf{pruning} as model compression techniques. While these methods reduce memory usage, they also significantly degrade performance, warranting further research.}. With the default batch size of 32, the total GPU memory consumption during inference is 12GB.

\section{Deployment Difficulty}

The RAG framework is encapsulated as a process class, facilitating easy loading and use, allowing for one-click deployment. We provide a Docker deployment script, with the Docker image size being approximately 28GB. We also offer API deployment scripts based on FastAPI\footnote{\url{https://fastapi.tiangolo.com/}} and a WebUI based on Streamlit\footnote{\url{https://streamlit.io/}}, making it convenient for use.

\section{Inference Latency}

\subsection{Standard Scheme}

\paragraph{Standard Time Delay} In the semi-final's standard scheme, we set the batch size for re-ranking to 32, with the inference latency for a question being 26 seconds, of which document sorting takes 6 seconds, and calling GLM4 twice takes 20 seconds.

\paragraph{Removing Answer Integration} By eliminating the answer integration step and directly returning the top 6 generated answers, only one call to GLM4 is needed, reducing the inference latency to 16 seconds.

\paragraph{Increasing Re-ranking Batch Size} Increasing the batch size to 256 increases the GPU memory usage but can reduce the inference latency to 24 seconds.

\paragraph{Full Process Acceleration Scheme} Beyond simple optimization strategies, we have also designed a full process acceleration scheme, which will be introduced in the following three subsections. This scheme aims to reduce time costs at each step. Due to the instability of GLM4 outputs, all experiments in this section terminate after the first generation of answers, without the final answer integration step, allowing for a more rigorous comparison of the impact of various acceleration methods on performance.

\subsection{BM25 Acceleration}

Since our retrieval stage relies heavily on BM25 for keyword matching, we introduced the bm25s~\cite{bm25s} library to optimize the speed of BM25 retrieval.

\begin{table}[htp]
\small
\centering
\begin{tabular}{ccc}
\toprule
 \textbf{Implementation} & \textbf{Time (s)} & \textbf{Accuracy} \\
\midrule
BM25Okapi & 17 & 94.49 \\
\textbf{BM25s} & 0.05 & 94.24 \\
\bottomrule
\end{tabular}
\caption{Effects of BM25 acceleration on the test set. Time represents the total search time for 103 questions related to BM25, and accuracy represents the evaluation score of the final generated answers.}
 \label{tab:eval}
\end{table}

\subsection{Reranker Acceleration}
We used the bge-reranker-v2-minicpm-layerwise model developed by the Zhejiang University's Institute for AI~\cite{chen2024bge} as the LLM Reranker. This model supports customization of the number of inference layers, allowing selection from 8-40 layers based on one's needs and resource constraints, thus reducing GPU memory overhead. In our preliminary experiments, we found that \textbf{28 layers} performed \textbf{slightly better than 40 layers}, with a difference of about 0.2 points, consistent with the empirical research conclusions given in the original repository. Therefore, both the preliminary and semi-final accuracy experiments utilized 28 layers.

However, since the Reranker is time-consuming in practical inference, we considered whether fewer layers could be used to speed up the process. Classic early-exit techniques in BERT, such as FastBERT~\cite{liu-etal-2020-fastbert} and DeeBERT~\cite{xin-etal-2020-deebert}, use information entropy exceeding a threshold as the condition for early exit, which is computationally intensive and results in unstable effects. Therefore, we designed a model early-exit algorithm based on maximum similarity selection, that is, for each query, we check if the softmax similarity output at the 12th layer in the first batch contains any values exceeding a certain threshold; if so, this query is inferred using just 12 layers, otherwise, 28 layers are used. We conducted an experiment using an A100 40G GPU to explore inference time, GPU memory usage, and accuracy at a batch size of 32, comparing different layers and early-exit methods. We randomly selected 10 queries and chose 192 text blocks for each, including 6 ground truth text blocks sorted using 28 layers in the complete RAG and 186 other random blocks. We predicted the sum of softmax scores of ground truth blocks relative to all blocks using various methods. Then, we assessed the similarity accuracy by dividing the predicted proportion by the proportion obtained with 28 layers, and compared the ranking accuracy of predicted ground truth with the 28-layer results, yielding the results shown in Table~\ref{tab:reranker}. It can be seen that our proposed model early-exit method, while reducing inference time by 33\%, is able to maintain ranking results consistent with those obtained using 28 layers directly, surpassing the entropy selection methods.

\begin{table}[htp]
\setlength{\tabcolsep}{1pt}
\centering
\begin{tabular}{cccc}
\toprule
 \textbf{Method} & \textbf{Time(s)} & \textbf{Similarity(\%)} & \textbf{Rank} \\
\midrule
8-layer & \textbf{1.67} & 73 & 2.5 \\
12-layer & 2.20 & \textbf{88} & 3.2 \\
20-layer & 3.58 & 86 & \textbf{4.0} \\
\midrule
28-layer & 5.25 & 100 & 6.0  \\
40-layer & 7.71 & 100 & 5.4 \\
\midrule
Maximum (0.1) & \textbf{2.59} & 90 & 3.7 \\
Maximum (0.2) & 3.55 & 96 & 4.5 \\
Maximum (0.4) & 4.57 & \textbf{97} & \textbf{5.4} \\
Entropy (0.2) & 2.74 & 89 & 3.4 \\
Entropy (0.4) & 3.37 & 91 & 3.6 \\
Entropy (0.6) & 4.01 & 91 & 4.0 \\
\bottomrule
\end{tabular}
\caption{Reranker Acceleration Experiment}
 \label{tab:reranker}
\end{table}

\subsection{Context Compression}

We designed a context compression method based on BM25 semantic similarity, which we call BM25-Extract. For each chunk, we first split it into sentences, then use BM25 to calculate the similarity between the query and each chunk, and finally add sentences to the list in order of decreasing similarity until a set compression rate is reached. The sentences are then concatenated in their original relative positions. We compared BM25-Extract with advanced context compression methods LLMLingua~\cite{jiang-etal-2023-llmlingua} and LongLLMLingua~\cite{jiang2023longllmlingua} as shown in Table~\ref{tab:compress}. Our method has advantages of no GPU memory usage, faster speed, and higher accuracy, making it evidently more effective for cost-sensitive operational maintenance tasks.

\begin{table}[htp]
\setlength{\tabcolsep}{3pt}
\tiny
\centering
\begin{tabular}{ccccc}
\toprule
 \textbf{Compression Algorithm} & \textbf{Compression Rate (\%)} & \textbf{Tokens Saved} & \textbf{Accuracy} & \textbf{Time (s)} \\
\midrule
Original Context & 100 & 0 & 94.49 & 9.30 \\
LLMLingua(0.5) & 62.80 & 143k & 83.44 & 10.47 \\
LongLLMLingua(0.5) & 62.80 & 143k & 80.86 & 10.52 \\
\textbf{BM25-Extract}(0.5) & \textbf{55.92} & \textbf{160k} & \textbf{86.48} & \textbf{7.70} \\
\textbf{BM25-Extract}(0.8) & \textbf{83.84} & \textbf{59k} & \textbf{89.00} & \textbf{8.12} \\
\bottomrule
\end{tabular}
\caption{Effects of context compression on the test set. Compression rate refers to the ratio of the length of the compressed prompt to the original prompt; tokens saved refers to the reduction in context string length divided by the empirical value of 1.6 to estimate the number of tokens saved; time refers to the average time per question from document retrieval to answer generation, including context compression and GLM4 generation time.}
 \label{tab:compress}
\end{table}

\section{Scalability}

\paragraph{Document Scalability} Our solution is primarily based on BM25 retrieval and Reranker re-ranking, requiring only processing of the latest documents, followed by re-segmentation and IDF value calculation. The entire process has a small time overhead and can be completed within 5 minutes.

\paragraph{User Scalability} Our solution has low GPU memory usage, and we have designed inference acceleration methods for various stages, allowing the use of specific optimization strategies depending on the user's scale. Even using a fully unaccelerated solution, a single 80GB GPU can support at least six RAG processes, returning answers to users within half a minute.

\section{Conclusion}
This paper presents EasyRAG, an accurate, lightweight, efficient, flexible, and scalable retrieval-augmented question-answering framework aimed at automated network operations.


\bibliography{custom}

\begin{thebibliography}{12}
\providecommand{\natexlab}[1]{#1}

\bibitem[{Chen et~al.(2024)Chen, Xiao, Zhang, Luo, Lian, and Liu}]{chen2024bge}
Jianlv Chen, Shitao Xiao, Peitian Zhang, Kun Luo, Defu Lian, and Zheng Liu. 2024.
\newblock \href {https://arxiv.org/abs/2402.03216} {Bge m3-embedding: Multi-lingual, multi-functionality, multi-granularity text embeddings through self-knowledge distillation}.
\newblock \emph{Preprint}, arXiv:2402.03216.

\bibitem[{Gao et~al.(2022)Gao, Ma, Lin, and Callan}]{gao2022precise}
Luyu Gao, Xueguang Ma, Jimmy Lin, and Jamie Callan. 2022.
\newblock Precise zero-shot dense retrieval without relevance labels.
\newblock \emph{arXiv preprint arXiv:2212.10496}.

\bibitem[{GLM et~al.(2024)GLM, Zeng, Xu, Wang, Zhang, Yin, Rojas, Feng, Zhao, Lai, Yu, Wang, Sun, Zhang, Cheng, Gui, Tang, Zhang, Li, Zhao, Wu, Zhong, Liu, Huang, Zhang, Zheng, Lu, Duan, Zhang, Cao, Yang, Tam, Zhao, Liu, Xia, Zhang, Gu, Lv, Liu, Liu, Yang, Song, Zhang, An, Xu, Niu, Yang, Li, Bai, Dong, Qi, Wang, Yang, Du, Hou, and Wang}]{glm2024chatglm}
Team GLM, Aohan Zeng, Bin Xu, Bowen Wang, Chenhui Zhang, Da~Yin, Diego Rojas, Guanyu Feng, Hanlin Zhao, Hanyu Lai, Hao Yu, Hongning Wang, Jiadai Sun, Jiajie Zhang, Jiale Cheng, Jiayi Gui, Jie Tang, Jing Zhang, Juanzi Li, Lei Zhao, Lindong Wu, Lucen Zhong, Mingdao Liu, Minlie Huang, Peng Zhang, Qinkai Zheng, Rui Lu, Shuaiqi Duan, Shudan Zhang, Shulin Cao, Shuxun Yang, Weng~Lam Tam, Wenyi Zhao, Xiao Liu, Xiao Xia, Xiaohan Zhang, Xiaotao Gu, Xin Lv, Xinghan Liu, Xinyi Liu, Xinyue Yang, Xixuan Song, Xunkai Zhang, Yifan An, Yifan Xu, Yilin Niu, Yuantao Yang, Yueyan Li, Yushi Bai, Yuxiao Dong, Zehan Qi, Zhaoyu Wang, Zhen Yang, Zhengxiao Du, Zhenyu Hou, and Zihan Wang. 2024.
\newblock \href {https://arxiv.org/abs/2406.12793} {Chatglm: A family of large language models from glm-130b to glm-4 all tools}.
\newblock \emph{Preprint}, arXiv:2406.12793.

\bibitem[{Jiang et~al.(2023{\natexlab{a}})Jiang, Wu, Lin, Yang, and Qiu}]{jiang-etal-2023-llmlingua}
Huiqiang Jiang, Qianhui Wu, Chin-Yew Lin, Yuqing Yang, and Lili Qiu. 2023{\natexlab{a}}.
\newblock \href {https://doi.org/10.18653/v1/2023.emnlp-main.825} {{LLML}ingua: Compressing prompts for accelerated inference of large language models}.
\newblock In \emph{Proceedings of the 2023 Conference on Empirical Methods in Natural Language Processing}, pages 13358--13376, Singapore. Association for Computational Linguistics.

\bibitem[{Jiang et~al.(2023{\natexlab{b}})Jiang, Wu, Luo, Li, Lin, Yang, and Qiu}]{jiang2023longllmlingua}
Huiqiang Jiang, Qianhui Wu, Xufang Luo, Dongsheng Li, Chin-Yew Lin, Yuqing Yang, and Lili Qiu. 2023{\natexlab{b}}.
\newblock Longllmlingua: Accelerating and enhancing llms in long context scenarios via prompt compression.
\newblock \emph{arXiv preprint arXiv:2310.06839}.

\bibitem[{Li et~al.(2023)Li, Zhang, Zhang, Long, Xie, and Zhang}]{li2023towards}
Zehan Li, Xin Zhang, Yanzhao Zhang, Dingkun Long, Pengjun Xie, and Meishan Zhang. 2023.
\newblock Towards general text embeddings with multi-stage contrastive learning.
\newblock \emph{arXiv preprint arXiv:2308.03281}.

\bibitem[{Liu et~al.(2020)Liu, Zhou, Wang, Zhao, Deng, and Ju}]{liu-etal-2020-fastbert}
Weijie Liu, Peng Zhou, Zhiruo Wang, Zhe Zhao, Haotang Deng, and Qi~Ju. 2020.
\newblock \href {https://doi.org/10.18653/v1/2020.acl-main.537} {{F}ast{BERT}: a self-distilling {BERT} with adaptive inference time}.
\newblock In \emph{Proceedings of the 58th Annual Meeting of the Association for Computational Linguistics}, pages 6035--6044, Online. Association for Computational Linguistics.

\bibitem[{Lù(2024)}]{bm25s}
Xing~Han Lù. 2024.
\newblock \href {https://arxiv.org/abs/2407.03618} {Bm25s: Orders of magnitude faster lexical search via eager sparse scoring}.
\newblock \emph{Preprint}, arXiv:2407.03618.

\bibitem[{Teo(2023)}]{costar}
Sheila Teo. 2023.
\newblock \href {https://medium.com/towards-data-science/how-i-won-singapores-gpt-4-prompt-engineering-competition-34c195a93d41} {How i won singapore’s gpt-4 prompt engineering competition}.

\bibitem[{Wang et~al.(2023)Wang, Yang, and Wei}]{wang2023query2doc}
Liang Wang, Nan Yang, and Furu Wei. 2023.
\newblock Query2doc: Query expansion with large language models.
\newblock \emph{arXiv preprint arXiv:2303.07678}.

\bibitem[{Wei et~al.(2022)Wei, Wang, Schuurmans, Bosma, Xia, Chi, Le, Zhou et~al.}]{wei2022chain}
Jason Wei, Xuezhi Wang, Dale Schuurmans, Maarten Bosma, Fei Xia, Ed~Chi, Quoc~V Le, Denny Zhou, et~al. 2022.
\newblock Chain-of-thought prompting elicits reasoning in large language models.
\newblock \emph{Advances in neural information processing systems}, 35:24824--24837.

\bibitem[{Xin et~al.(2020)Xin, Tang, Lee, Yu, and Lin}]{xin-etal-2020-deebert}
Ji~Xin, Raphael Tang, Jaejun Lee, Yaoliang Yu, and Jimmy Lin. 2020.
\newblock \href {https://doi.org/10.18653/v1/2020.acl-main.204} {{D}ee{BERT}: Dynamic early exiting for accelerating {BERT} inference}.
\newblock In \emph{Proceedings of the 58th Annual Meeting of the Association for Computational Linguistics}, pages 2246--2251, Online. Association for Computational Linguistics.

\end{thebibliography}

\appendix

\section{Question-and-Answer Prompt Templates}

\subsection{Markdown Format Question-and-Answer Template}
\begin{tcolorbox}\label{sec:md_temp}
\#\# Objective\\

Please, based on the information from {k} private domain documents about 5G operational maintenance, answer the given question.\\

\#\# Requirements\\

1. You may itemize your answer; be as detailed and specific as possible.\\
2. Do not merely repeat information from the context.\\
3. Do not use your own knowledge; rely solely on the content from the context documents.\\

\#\# Context\\

\{{\color{blue}context\_str}\}\\

\#\# Question\\

\{{\color{blue}query\_str}\}\\

\#\# Answer\\
\end{tcolorbox}

\subsection{Chain of Thought Question-and-Answer Template}
\begin{tcolorbox}\label{sec:cot_temp}
Context information as follows:

-\--\--\--\--\--\--\--\--\--

\{{\color{blue}context\_str}\}

-\--\--\--\--\--\--\--\--\--\\

Please answer the following question based on the context information rather than your own knowledge. Think step by step, first provide an analysis process, then generate an answer:\\

\{{\color{blue}query\_str}\}\\

Answer:
\end{tcolorbox}

\subsection{Focused Question-and-Answer Template}
\begin{tcolorbox}\label{sec:lean_temp}
Context information as follows:

-\--\--\--\--\--\--\--\--\--

\{{\color{blue}context\_str}\}

-\--\--\--\--\--\--\--\--\--\\

Please answer the following question based on the context information rather than your own knowledge. You may itemize your answer. Document 0's content is particularly important, consider it carefully. If the context does not contain relevant knowledge, you may respond with 'uncertain'. Do not simply restate the context information:\\

\{{\color{blue}query\_str}\}\\

Answer:
\end{tcolorbox}

\section{Answer Integration Template}

\begin{tcolorbox}\label{sec:merge_temp}
Context:

-\--\--\--\--\--\--\--\--\--

\{{\color{blue}top1\_content\_str}\}

-\--\--\--\--\--\--\--\--\--\\

You will see a question and a corresponding reference answer\\

Please, based on the context knowledge and not your own knowledge, supplement the reference answer to make it more complete in addressing the question\\

Please note, strictly retain every character of the reference answer and reasonably integrate your supplement with the reference answer to produce a longer, more complete answer containing more terms and itemization\\

Question:

\{{\color{blue}query\_str}\}\\

Reference answer:

\{{\color{blue}answer\_str}\}\\

New answer:

\end{tcolorbox}

\end{document}